\documentclass{article}

\usepackage{arxiv}

\usepackage[utf8]{inputenc} 
\usepackage[T1]{fontenc}    
\usepackage{hyperref}       
\usepackage{url}            
\usepackage{booktabs}       
\usepackage{amsfonts}       
\usepackage{nicefrac}       
\usepackage{microtype}      
\usepackage{lipsum}
\usepackage{graphicx}
\graphicspath{ {./images/} }
\usepackage{amsmath}
\usepackage{wrapfig}

\usepackage{cleveref}
\usepackage{adjustbox}
\usepackage{xcolor}
\usepackage{natbib}
\usepackage{graphicx}
\usepackage{subcaption}
\usepackage{pgfplots}
\usepackage{mdframed}
\usepackage{cleveref}

\definecolor{ForestGray}{HTML}{004529}

\colorlet{definitioncolour}{ForestGray!10!white}

\mdfdefinestyle{fancyenv}{
    skipabove=0.8\baselineskip,
    skipbelow=0.\baselineskip,
    innertopmargin=8pt,
    innerbottommargin=8pt,
    linewidth=2pt,
    topline=true,
    frametitleaboveskip=\dimexpr-\ht\strutbox\relax,
    nobreak=true,
}

\newcounter{definition}

\definecolor{codegreen}{rgb}{0,0.6,0}
\definecolor{codegray}{rgb}{0.5,0.5,0.5}
\definecolor{codepurple}{rgb}{0.58,0,0.82}
\definecolor{backcolour}{rgb}{0.95,0.95,0.92}
\definecolor{jcred}{HTML}{e31a1c}
\definecolor{jcgreen}{HTML}{33a02c}
\definecolor{jcblue}{HTML}{1f78b4}
\definecolor{jcorange}{HTML}{ff7f00}
\definecolor{jcpurple}{HTML}{6a3d9a}
\definecolor{jclightred}{HTML}{fb8072}
\definecolor{jclightgreen}{HTML}{b3de69}
\definecolor{jclightblue}{HTML}{80b1d3}
\definecolor{jclightorange}{HTML}{fdb462}
\definecolor{jclightpurple}{HTML}{bebada}
\definecolor{jcredl}{HTML}{fb8072}
\definecolor{jcgreenl}{HTML}{b3de69}
\definecolor{jcbluel}{HTML}{80b1d3}
\definecolor{jcorangel}{HTML}{fdb462}
\definecolor{jcpurplel}{HTML}{bebada}
\definecolor{jcbluem}{HTML}{488bb8}

\title{Heterogeneous Computing: The Key to Powering the Future of AI Agent Inference}

\author{
 Aaron Zhao \\
  Imperial College London\\
  \texttt{a.zhao@imperial.ac.uk} \\
  \And
 Junyi Liu \\
  Microsoft Research\\
  \texttt{junyi.liu@microsoft.com}
  \texttt{} \\
}
\date{} 
\begin{document}
\maketitle

\section{Introduction}

The rise of AI agents has significantly reshaped the landscape of computer systems.
There are gigawatt-scale data centers to be built in the coming years, primarily to support AI workloads \citep{gigawatt}.
We are moving toward an inference-heavy future -- reports have shown that AI agents are expected to power many existing and future applications, and running inference (just the forward pass) for these agents will likely become the dominating workloads in future AI data centers \citep{mckinsey2025nextshifts,morganstanley2025inference}.


Hardware heterogeneity has become a key part of today's cloud computing.
Starting from optimizing CPU-oriented workloads, we have seen the adoption of networking accelerators like SmartNICs and parallel compute units like GPUs.
These accelerators greatly improve both system performance and efficiency by offloading IO- and compute-intensive tasks.
Since the release of ChatGPT, AI workloads have been predominantly optimized for the GPU-centric infrastructure.
Today, a predominant belief is that more compute specialization will be needed for AI agent inference, and we see this trend has started in the inference chips like Nvidia's Rubin CPX \citep{nvidia2024rubincpx}.
In fact, hardware heterogeneity is further spreading beyond the compute accelerators.
Scalable and efficient inference for large-scale AI agents requires wider and spread-out heterogeneity at the system level ‐ across compute, networking, and memory.

AI agent inference, when deployed across different scenarios, can now exhibit substantial variation in compute to memory bandwidth (FLOPs/bandwidth) and compute to memory capacity (FLOPs/capacity) ratios. This results from a combination of different \textbf{agentic stack use cases} (e.g., computer use, chatbot), \textbf{base model architectures} (e.g., different attentions employed, dense or sparse architectures), and distinct system and model-level \textbf{optimizations} (e.g., quantization, prefill-decode disaggregation). To satisfy such workload diversity, future scaling imposes unprecedented pressures on accelerators, interconnects, and memory systems.
We argue that system‑level heterogeneity will, in turn, require cohesive integration at the datacenter scale.

\section{Operational Intensity and Capacity Footprint}

We define two key metrics -- related to the capacity and bandwidth requirements on the memory side -- to better characterize the inference process of a given AI agent.:
\begin{itemize}
\item \textbf{Operational Intensity (OI)}: The number of operations performed per byte of data moved from DRAM, this is a classic metric widely used in the roofline models.
\item \textbf{Capacity Footprint (CF)}: The number of bytes needed per agent request in DRAM for LLM generation. The product of batch size and CF states the capacity requirement on the DRAM.
\end{itemize}
Consider a simple matrix multiplication $\mathbf{Y} = \mathbf{W}\mathbf{X}$, where $\mathbf{W} \in \mathcal{R}^{m \times d}$ and $\mathbf{X} \in \mathcal{R}^{d \times L}$, assuming we have a sequence length $L$ and hidden dimension $d$.
This means that the given OI is the total number of operations divided by the amount of memory transactions needed (loading $\mathbf{W}$ and $\mathbf{X}$, and writing out $\mathbf{Y}$). This gives an operations per bandwidth of:
$\frac{2mdL}{md + dL + mL}$.
Similarly, the capacity requirement is to store only $\mathbf{W}$ in most AI inference workloads. In LLM inference, both $\mathbf{Y}$ and $\mathbf{X}$ are activation values transferred between consecutive transformer layers, which can be kept in on-chip SRAM or temporarily buffered in DRAM.
So the CF is simply $\frac{md}{B}$ in this case, where $B$ is the batch size \footnote[1]{In this case, only $\mathbf{W}$ is considered as DRAM storage requirements, because $\mathbf{X}$ and $\mathbf{Y}$ are temporary values and are also vectors which is much smaller. In the KV caching case, we would then modify the equation to consider storing $\mathbf{K}, \mathbf{V} \in \mathcal{R}^{d \times L}$ for each request. This will change CF to be $2dL + \frac{md}{B}$. }.

\begin{figure}[!th]
    \centering
    \includegraphics[width=.9\linewidth]{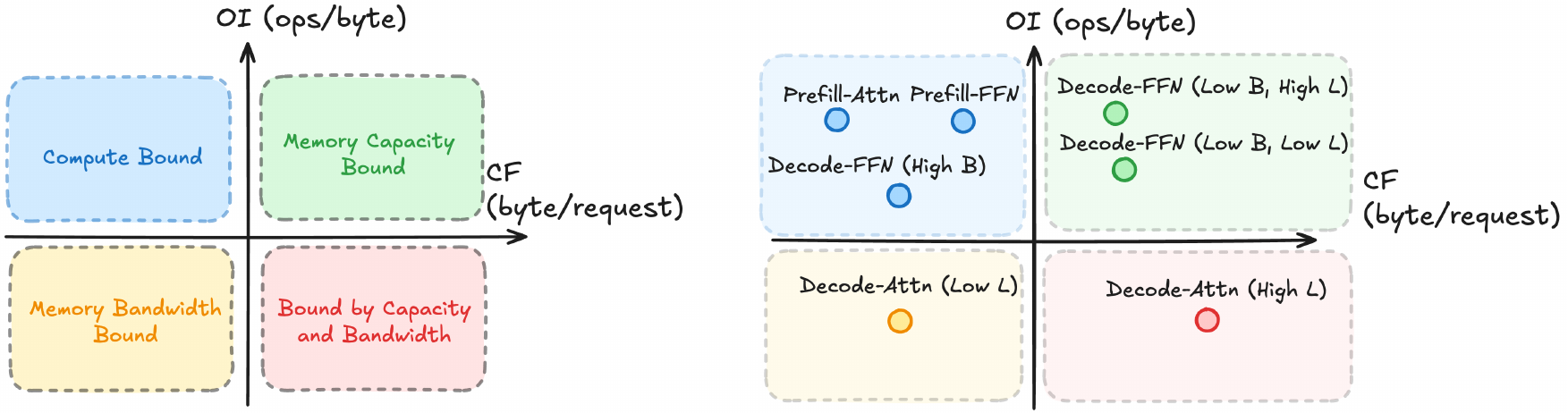}
    \caption{The traditional roofline model only addresses the leftmost sections (the blue and yellow areas). High MFU indicates the compute-bound region (blue), while high MBU represents the memory bandwidth-bound region (yellow). However, both the roofline model and the concepts of MBU and MFU fail to clearly illustrate under-utilization that arises from memory capacity limitations. We also map actual workloads to these regions, for instance, Decode-FFN (low B, high L) means decode-time FFN block in transformers at low batch size (B) and long sequence length (L).}
    \label{fig:intro}
\end{figure}
\subsection{Addressing the limitations of existing theoretical models}

These metrics connect directly to the classic roofline model \citep{williams2009roofline} and also the popular Model FLOPs Utilization (MFU) and Memory Bandwidth Utilization (MBU) descriptors used for ML systems \citep{databricks2024llminference}. However, OI and CF together provide a more fine-grained and complete picture for the AI agent inference system.

As shown in \Cref{fig:intro},
the Compute per Bandwidth metric essentially measures operational intensity: if this value is low, the workload execution is limited by the available memory bandwidth—the so-called "memory wall" phenomenon described in current literature \cite{gholami2024ai}—rather than by the available compute resources.
This is also the main focus of the classic roofline model \citep{williams2009roofline}, which explains that a workload can be either compute bound or memory bandwidth bound. The same reasoning can be applied on using the MBU and MFU descriptors in an AI agent inference system: when MBU is high and MFU is low, the system is considered memory bandwidth bound; when MFU is high and MBU is low, the system is then compute bound.

However, in practice, a limitation arises: when serving LLM token generation, one can often observe low FLOPs even before reaching the memory bandwidth roof. Translating this to the MBU and MFU descriptors, it means both MBU and MFU can be low at the same time. The whole system can be limited by other factors that these existing theoretical models failed to capture, and the missing dimension is the memory capacity \footnote[2]{For memory here, we refer to off-compute-die memory, such as DRAM and HBM.}. This is also referred as the ``memory capacity wall" \citep{wu2025combating}.
Essentially, the existing roofline model does not adequately capture the blue and yellow quadrants shown in \Cref{fig:intro}, which turn out to be an extremely important limiting factor in AI agent inference.

It should be noted that there is another potential limiting factor: chip interconnect communication bandwidth. This is normally a constraint from the networking stack.
However, this is the subsequent bottleneck in AI inference after the constraints set by memory bandwidth and capacity (such as DDR and HBM). For instance, transferring typical intermediate activation tensors rarely exceeds the bandwidth of the scale-up domain network, such as NVLink domain. Nevertheless, as we continue to scale for larger models using extensive Tensor Parallelism and Expert Parallelism, we are quickly approaching this limitation. Additionally, this bottleneck becomes a more significant factor during AI training, since gradients must be gathered and updated weights broadcast, though this aspect is outside the scope of this discussion.


\subsection{What affects OI and CF?}

\begin{figure}[t]
\centering
\includegraphics[width=\linewidth]{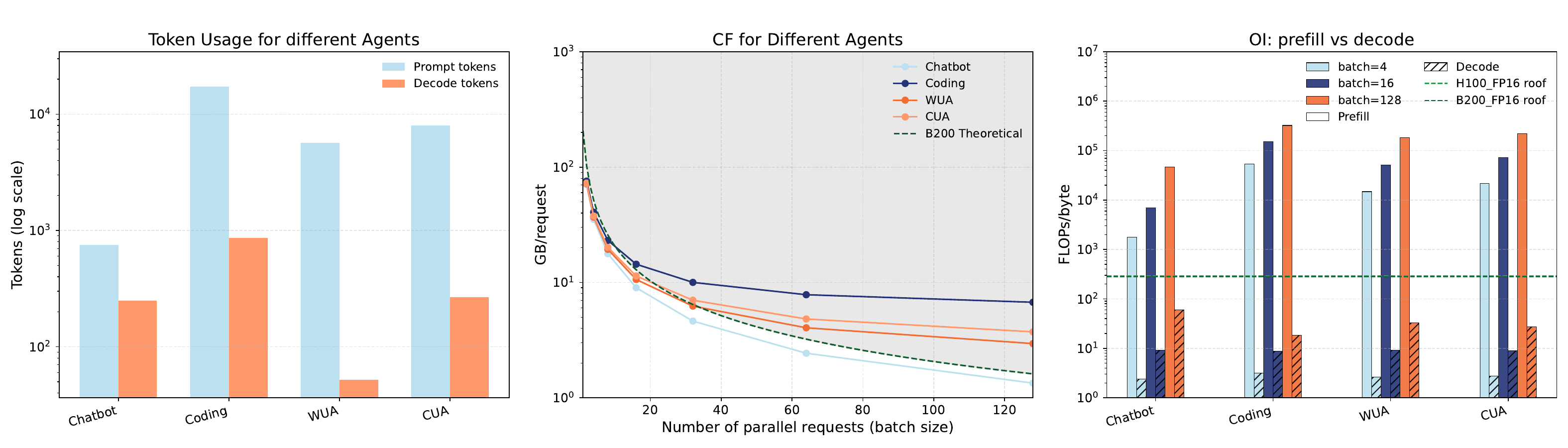}
\caption{Prefill and decode token usage, as well as CF and OI characteristics, for different agents (Chatbot, Coding, Web-use, and Computer-use) using a dense LLaMA3-70B model.}
\label{fig:agent}
\end{figure}

Several key factors related to usage scenarios, model design, and optimizations can significantly influence OI and CF. Their impact can often be substantial enough to shift workload characteristics across the different quadrants shown in \Cref{fig:intro}, thereby greatly affecting how one should design a system to serve models efficiently.

\subsubsection{Different agentic workflows}
Many existing hardware accelerators target chatbots as the primary usage scenario. However, the input characteristics can vary significantly when considering different types of agents, even when using the same base model.
In \Cref{fig:agent}, we present token utilization characteristics for a Chatbot \citep{anthropic2025pricing}, Coding agent \citep{fan2025sweeffi}, Web-use agent \citep{wang2025awm}, and Computer-use agent \citep{agashe2025agents2}. We fit a LLaMA-70B model with these input-output prompt characteristics and illustrate its CF and OI behaviors. As shown, different agentic setups can lead to noticeable variations in CF and OI.
It is also worth noting that, in such agentic workflows, the ``snowballing effect" is commonly observed -- context length rapidly increases to unmanageable levels. 
This phenomenon is especially prevalent in coding, web-use, and computer-use agents. Intuitively, these tasks (e.g., coding or GUI interactions) require multiple environment-agent exchanges; studies report an average of 20-30 interactions per task \citep{agashe2025agents2}. Taking a coding agent as an example, the context length can quickly grow to 300K tokens with some cases exceeding 1M tokens, creating immense pressure on memory capacity and resulting in low OI due to the heavy KV cache loading required. 
The middle of \Cref{fig:agent} shows how CF for most agentic workloads can exceed the capacity of a single B200 card (the gray area), while the rightmost part illustrates that, due to the need to load a large amount of KV cache values, the OI during the decoding phase is extremely low. As a result, the hardware spends most of its time loading from and writing to DRAM rather than performing useful computation. This creates a dilemma: more cards are needed due to the CF limit, but this is inefficient because the OI remains far below the best operating value, and adding more cards does not improve it.

\subsection{Base model variations and optimizations}

\begin{figure}
    \centering
    \includegraphics[width=.45\linewidth]{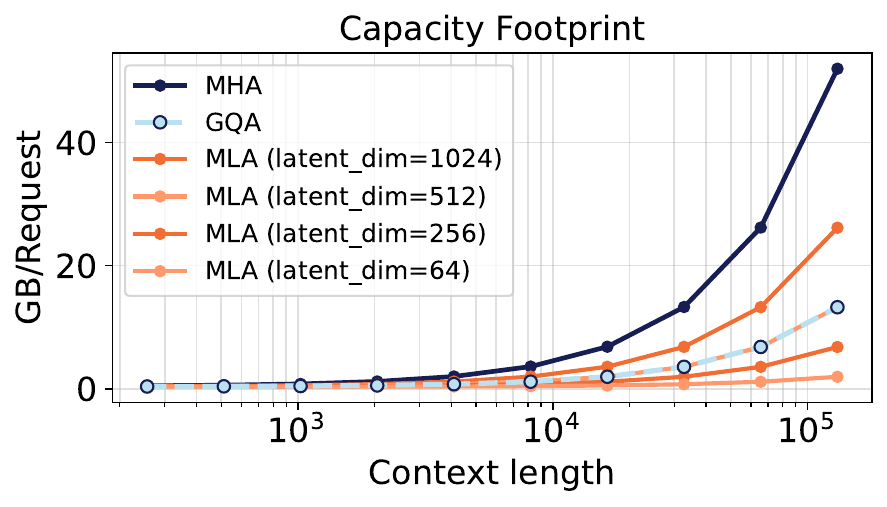}
    \caption{We compare Multi-head Attention (MHA), Grouped Query Attention (GQA), and Multi-head Latent Attention (MLA) using a 48-layer model with a hidden dimension of 2048 and 32 attention heads, all operating at 16-bit precision.}
    \label{fig:atten}
\end{figure}

\begin{figure}
    \centering
    \includegraphics[width=\linewidth]{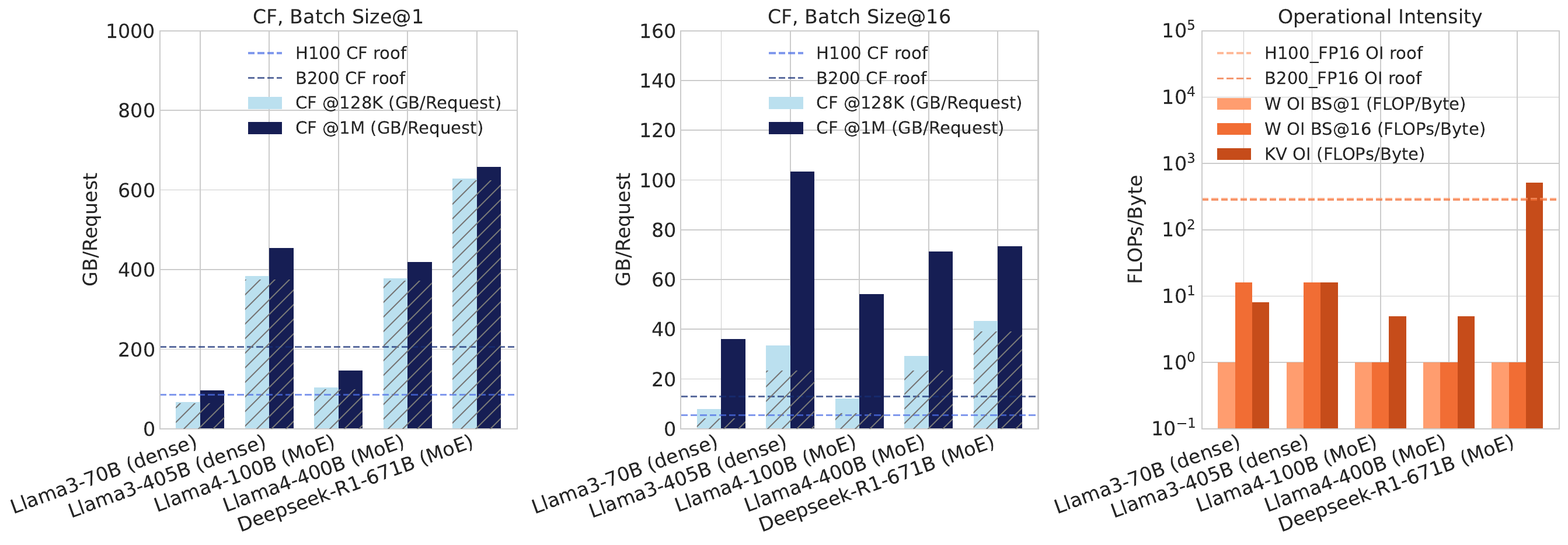}
    \caption{Capacity Footprint (for batch sizes 1 and 16) and Decode time OI, comparing dense models and sparse MoE models. The shaded region represents model weights.}
    \label{fig:moe}
\end{figure}

We now see a wide range of base architectures with various optimizations, some of which may affect how the base model performs. Therefore, we identify a set of possible optimizations that can undoubtedly have a significant impact on CF and OI values.
\begin{itemize}
    \item Different attention mechanisms: As shown in \Cref{fig:atten}, these attention types in standard models (GPT \citep{brown2020language}, LLaMA \citep{touvron2023llama}, DeepSeek \citep{liu2024deepseek}, Qwen \citep{yang2025qwen3}) exhibit drastically different behaviors in terms of capacity footprint, especially at higher context lengths. While it is feasible to serve models with efficient attention schemes (e.g., MLA) at short context lengths using small HBMs, such hardware would likely struggle at larger context lengths. Notably, there is also a rise in other attention schemes that claim to be linear in both compute and memory usage \cite{wang2020linformer}, although their effectiveness at large scale has yet to be fully validated.
    \item Model Sparsity: The introduction of mixture-of-experts (MoE) has enabled models to maintain parameter scaling \citep{shazeer2017outrageously}. MoE effectively reduces OI by activating only a small subset of the total weights for compute. As shown in \Cref{fig:moe}, CF of MoE models is sensitive to both model weights and long-context KV cache. Additionally, when considering OI, MoE models are significantly more memory-bound.
    Additionally, model sparsity is being actively explored on token sequence dimension. KV cache compression and sparse attention techniques have the potential to greatly reduce KV cache read size at inference time.
    \item Quantization: Quantization remains one of the most effective methods for reducing not only compute requirements but also memory capacity and bandwidth needs, thereby directly lowering CF. Nvidia’s release of NVFP4 and GPT-OSS’s native support for this format further confirm this trend \citep{agarwal2025gpt}. However, with 4-bit inference being widely explored from the edge side, it is debatable how much further improvement is possible. One potential direction is asymmetric quantization \citep{zhang2024qera}, though even that may at best allow us to reach the 2-bit era. 
\end{itemize}

\subsection{System-level implications}

Based on our observations of how different design choices in AI agents can drastically change inference behaviors, we can extract several interesting implications:

\begin{enumerate}
    \item Disaggregated serving should be the default approach—as shown on the rightmost side of \Cref{fig:agent}, prefill and decode phases exhibit drastically different operational intensity. This suggests that we should design specialized accelerators for these two phases independently. More importantly, the middle plot of \Cref{fig:agent} further emphasizes that large memory capacity is critical to satisfy the rapid growth of diverse agentic workloads.
    \item Agentic AI requires more prefill tokens and supports for larger context lengths. As shown in \Cref{fig:agent}, the complexity of agentic tool definitions and the nature of these tasks result in more prefill tokens being used. Additionally, multi-step and multi-round interactions demand longer context lengths.
    \item Memory capacity and bandwidth are now the two critical bottlenecks for long-context inference. Although efficient attention schemes (\Cref{fig:atten}) and MoE architectures (\Cref{fig:moe}) helped, the KV cache has now become the main challenge for both memory capacity and bandwidth. Efficient system-level parallelism strategies (e.g., pipeline parallelism), along with heterogeneous compute architectures, memory systems, and networking stacks, can help address this major challenge.
\end{enumerate}

\section{The heterogeneity panacea}

Today, advanced packaging and DRAM die stacking are barely driving the performance scaling of AI systems.
There is still a planned roadmap for packaging more reticle-sized compute dies and increasing the stacking density of HBMs. 
Similar to Moore's law, the speed of adopting new packaging technologies may already show signs of slowing down.
Meanwhile, the scaling of compute FLOPs, memory bandwidth, capacity, and networking speed has been highly asymmetric.
Compute FLOPs is increasing much faster than memory and interconnect.
In particular, the demand of memory has never been stronger and more diverse than before. 
Sticking to homogeneous chips in AI datacenters will amplify mismatched scaling and will eventually create severe cost/performance bottlenecks.

More than ever, we need ways to scale compute beyond the package boundary.
The pursuit of advanced optical IO technologies will reshape the heterogeneous architecture of AI systems, which drives long-term system scaling.
With future optical IOs that will potentially provide D2D-scale bandwidth and <1pJ/bit energy efficiency\citep{benyahya2025mosaic}, disaggregating compute and memory will become feasible to reach the next level of system efficiency by introducing heterogeneity.
Compute chips with different FLOPs will be interconnected together, and they will access different memory devices at the system level \citep{canakci2025good}.
The scale-up network domain will keep expanding with wider adoption of heterogeneous switching (e.g., optical circuit switching in TPUs\cite{jouppi2025ironwood}).

According to the emerging OI and CF in future agentic AI workloads, such systems will be able to adapt their selection of die-level FLOPs and memory devices like SRAMs, HBMs, LPDDRs, and SSDs. 
Hardware-level adaptation can be applied after the system is deployed in the data center.
If OI and CF of agentic workloads shift substantially, a heterogeneous, fully disaggregated AI system can be rebalanced within a single hardware generation rather than waiting for new hardware — something today’s homogeneity-centric AI systems cannot do.



\section{Where Agentic AI inference systems are headed: Hypotheses}

\noindent
\textbf{Hypothesis 1:} \textit{The future of AI serving lies in disaggregated architectures tailored for distinct prefill and decode phases.}

NVIDIA's roadmap already includes prefill-only chips, such as the upcoming Rubin CPX. However, we anticipate the emergence of \textbf{more than two types of inference accelerators} within a single inference system driven by the needs to support different agents and different input modalities. These accelerators may feature varying memory capacity-to-compute or memory bandwidth-to-compute ratios. Furthermore, we expect to see support for different sets arithmetic precisions for prefill and decode computations. In particular, we hypothesize that certain arithmetic formats may be better suited for prefill, while other arithmetic types could be more advantageous for decode.

\noindent
\textbf{Hypothesis 2:} \textit{Agents will be co-designed with the serving hardware through pre-training and post-training adaptations.} 

We hypothesize that models will first be trained on large-scale infrastructure and then distilled into variants optimized for efficient inference. Such distillation can occur during both the pre-training and post-training stages. Old techniques such as hardware-aware network architecture search could be revamped to integrate to such distillation process.
This co-design will account for factors such as network topology, total available HBM capacity and bandwidth, and the arithmetic formats supported by the target deployment hardware. Without identical infrastructure, it will be challenging for third parties to even achieve the same level of inference efficiency. 

\noindent
\textbf{Hypothesis 3:} \textit{A new training/inference paradigm emerges and completely transforms the hardware landscape.}

Whether this is good or bad news remains to be seen, but we have already witnessed the emergence of different training and inference paradigms, such as diffusion models and state space models. It is entirely possible that these new models could redefine the system stack that currently exists or is being developed. Alternatively, it is also likely that such models could be distilled from large auto-regressive models we have today and applied to specific scenarios, such as edge computing.

\noindent
\textbf{Hypothesis 4:} \textit{The increase of scale-up networking domain size will allow more compute hardware heterogeneity.}
Optical IOs based on co-packaged optics will keep expanding the scale-up networking domain. 
Beyond dedicated compute chips for prefill and decode phases, there will be growing demand for compute chips optimized for low latency and physical-world interaction. 

\noindent
\textbf{Hypothesis 5:} \textit{High-bandwidth and large-capacity memory disaggregation will sustain the growth of agent memory usage}
Diverse agent-level memory has been widely developed for complex agentic workflows.
Agents will need both short-term and long-term working memory that can be backed by various data stores such as KV-cache stores, knowledge databases and file systems. 
System‑level disaggregated memory will enable continued scaling of its bandwidth and capacity for AI agent workloads.


\section{Conclusion}
In summary, by introducing Operational Intensity (OI) and Capacity Footprint (CF), we present a fine-grained framework for understanding various bottlenecks in agentic AI agent inference workloads. These metrics expose the limitations of classic roofline models and commonly used descriptors like MFU and MBU, especially in highlighting the impact of memory capacity constraints that arise in long-context, agentic tasks. Our analysis demonstrates that model architecture choices, workload scenarios, and optimization techniques can shift the balance between OI and CF, sometimes moving workloads into fundamentally different system bottleneck regimes. These insights underscore the need for rethinking hardware and system design -- pointing to disaggregated compute, heterogeneous architectures, and workload-aware co-design as essential directions to keep scaling AI agent systems, break through existing bottlenecks, and unlock new levels of efficiency and capability.

\bibliographystyle{abbrv}


\bibliography{references}
\end{document}